\documentclass[runningheads, orivec]{llncs}
\usepackage{cite}
\usepackage{times}
\usepackage{epsfig}
\usepackage{graphicx}
\usepackage{amsmath}
\usepackage{amssymb}
\usepackage{multirow}
\usepackage{mathtools}
\usepackage[pagebackref=true,breaklinks=true,colorlinks,bookmarks=false]{hyperref}

\usepackage{array}
\newcolumntype{H}{>{\setbox0=\hbox\bgroup}c<{\egroup}@{}}

\newcommand{\red}[1]{\textcolor{red}{#1}}
\newcommand{\one}[1]{\textbf{#1}}

\begin{document}
\title{Recursively Refined R-CNN: \\Instance Segmentation with Self-RoI Rebalancing}
%
%
\author{Leonardo Rossi\inst{1}\orcidID{0000-0002-9316-595X} \and
Akbar Karimi\inst{1}\orcidID{0000-0002-5132-2435} \and
Andrea Prati\inst{1}\orcidID{0000-0002-1211-529X}}
\authorrunning{L. Rossi et al.}
%
\institute{IMP Lab - D.I.A. - University of Parma - Parma, Italy\\
\email{\{leonardo.rossi, akbar.karimi, andrea.prati\}@unipr.it}\\
\url{implab.ce.unipr.it/}}
\maketitle              
\begin{abstract}
Within the field of instance segmentation, most of the state-of-the-art deep learning networks rely nowadays on cascade architectures \cite{Cai_2018_CVPR}, where multiple object detectors are trained sequentially, re-sampling the ground truth at each step.
This offers a solution to the problem of exponentially vanishing positive samples.
However, it also translates into an increase in network complexity in terms of the number of parameters.
To address this issue, we propose Recursively Refined R-CNN ($R^3$-CNN) which avoids duplicates by introducing a loop mechanism instead. At the same time, it achieves a quality boost using a recursive re-sampling technique, where a specific IoU quality is utilized in each recursion to eventually equally cover the positive spectrum.
Our experiments highlight the specific encoding of the loop mechanism in the weights, requiring its usage at inference time.
The $R^3$-CNN architecture is able to surpass the recently proposed HTC \cite{chen2019hybrid} model, while reducing the number of parameters significantly.
Experiments on COCO \textit{minival} 2017 dataset show performance boost independently from the utilized baseline model.
The code is available online at \url{https://github.com/IMPLabUniPr/mmdetection/tree/r3_cnn}.

\end{abstract}

\section{Introduction}
\label{sec:introduction}

Computer vision is a field of continuous experimentation, where new and better performing algorithms are developed every day and are able to operate in environments with increasingly extreme conditions.
In particular, object detection, and instance segmentation as its narrower extension, offers complex challenges which are utilized in various applications, including medical diagnostics \cite{chen2017dcan}, autonomous driving \cite{huang2019robust}, visual product search \cite{liu2015matching}, and many others.
All these applications demand high-performing systems in terms of prediction quality, as well as low memory usage.
Therefore, a desirable architecture is as light as possible regarding the parameter count since it reduces the search space and enhances generalization, while retaining high-quality detection and segmentation performance.

However, often these two goals are conflicting.
The $R^3$-CNN architecture and the corresponding training mechanism that we propose present a trade-off between these two conflicting goals.
We show that our model is able to obtain the same performance of complex networks (such as HTC \cite{chen2019hybrid}) with a network as light as Mask R-CNN \cite{he2017mask}. 

The accuracy of instance segmentation systems is strongly based on the concept of intersection over union (IoU), which is used to identify the detection precision with respect to the ground truth.
The higher this value is, the more accurate and the less noisy the predictions are.
However, by increasing the IoU threshold, a problem called \emph{exponentially vanishing positive samples} \cite{Cai_2018_CVPR} (EVPS) is also introduced, meaning that it can give rise to the problem of good proposals scarcity compared to low-quality ones.
This usually leads to a training that is excessively biased towards low-quality predictions.
In order to solve this issue, Cascade R-CNN \cite{Cai_2018_CVPR} first, and its descendant HTC later, introduced a cascade mechanism where multiple object detectors are trained sequentially in order to take advantage of the previous one and to increase the prediction quality gradually.
This means that each stage performs two tasks: first, the detector is training itself, and, then, it is also devoted to identifying the region proposals for subsequent stages.
Unfortunately, this also translates into an increase in network complexity in terms of the number of parameters.

In this work, we propose a new way to balance positive samples by exploiting the re-sampling technique, introduced by the cascade models.
Our proposed technique generates new proposals with a pre-selected IoU quality in order to equally cover all IoU values.
We carry out an extensive ablation study and compare our results with the state of the art in order to demonstrate the advantages of the proposed solution and its applicability to different existing architectures. 

The main contributions of this paper are the following:
\begin{itemize}
	\item An effective solution to deal with the EVPS problem with a single-detector model, rebalancing the proposals with respect to the IoU thresholds through a recursive re-sampling mechanism. This mechanism has the goal of eventually feeding the network with an equal distribution of samples.
	\item An exhaustive ablation study on all the components of our $R^3$-CNN architecture in order to evaluate how the performance is affected by each component.
	\item Our $R^3$-CNN is introduced into major state-of-the-art models to demonstrate that it boosts the performance independently from the baseline model used.
\end{itemize}
\section{Related Works}
\label{sec:related}

\noindent\textbf{Multi-stage Detection/Instance Segmentation}.
The early works on object detection and instance segmentation were based on the assumption that single-stage end-to-end networks are sufficient to recognize and segment the objects. For instance, YOLO network \cite{redmon2016you} optimizes localization and classification in one step.
Starting with the R-CNN network \cite{girshick2014rich}, the idea of a two-stage architecture was introduced, where, in the first stage, a network called RPN (Region Proposal Network) analyzes the whole image and identifies the regions where the probability of finding an object is high. In the second stage, another network performs a more refined analysis on each single region.
After this seminal work, others have further refined this idea.
The Cascade R-CNN architecture \cite{Cai_2018_CVPR} uses multiple bounding-box heads connected sequentially, where each one refines the proposals produced by the previous one.
The minimum IoU required for positive examples is increased at each stage, taking into account a different set of proposals.
Other studies \cite{vu2019cascade, wang2019region, zhong2020cascade} introduced a similar cascade concept, but applied to the RPN network, where
multiple RPNs are sequenced and the results from the previous stage are fed into the next stage.
Our work is inspired by HTC network \cite{chen2019hybrid}, which introduces a particular cascade operation also on the mask extraction modules.
However, all these multi-stage networks are quite complex in terms of the number of parameters.

\noindent\textbf{IoU distribution imbalance}.
Authors in \cite{oksuz2020imbalance} describe the problem as a skewed IoU distribution observed in bounding boxes used in training and evaluation.
In \cite{shrivastava2016training}, the authors highlight the significant imbalance between background and foreground RoI examples and present a hard example mining algorithm to easily select the most significant ones.
While in their case the aim is balancing the background (negative) and the foreground (positive) RoIs, in our work the primary goal is to balance RoIs across the entire positive spectrum of the IoU.
In \cite{pang2019libra}, an IoU-balanced sampling technique is proposed to mine hard examples.
However, the sampling always takes place on the results of the RPN which, as we will see, is not very optimized to provide high-quality RoIs.
In our case, we apply re-sampling to the detector itself, which has, on average, a much higher probability of returning more significant RoIs.
In \cite{cheng2018revisiting}, the sources of false positives are analyzed and an extra independent classification head is introduced to be plugged into the original architecture to reduce hard false positives.
In \cite{zhu2019iou}, the authors introduce a new IoU-prediction branch which supports classification and localization.
They propose to manually generate samples around ground truths instead of using RPN for localization and IoU prediction branches in training.
In \cite{Cai_2018_CVPR, chen2019hybrid}, overfitting due to EVPS problem for large thresholds is addressed using multiple detectors connected sequentially.
They re-sample the ground truth in a sequential manner to progressively improve hypothesis quality.
Unlike them, we tackle the problem with a single detector and a single segmentation head.
In \cite{oksuz2020generating}, they offer an interpretation similar to ours about the fact that IoU imbalance has an adverse effect on performance. However, while they use an algorithm to systematically generate the RoIs with the chosen quality, we rely only on the capabilities of the detector itself.

\section{Recursively Refined R-CNN}
\label{sec:theory}

In this section, first we briefly introduce the idea behind multi-stage processing. Then we describe our $R^3$-CNN architecture with its evolution from a sequential to a recursive pipeline, which offers a change of perspective on training.

\begin{figure*}[t]
	\begin{center}
		\includegraphics[trim=20 340 0 130, clip, width=1.1\linewidth]{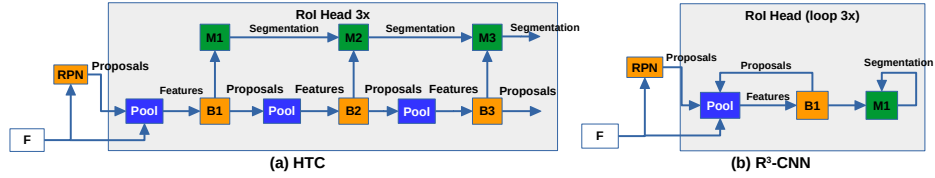}
	\end{center}
	\caption{Network design. (a) HTC: a multi-stage network which trains each head in a cascade fashion. (b) $R^3$-CNN: our architecture which introduces a loop mechanism to self-train the heads.}
	\label{fig:recref-arch}
\end{figure*}

As shown in Fig. \ref{fig:recref-arch} (a), the HTC (Hybrid Task Cascade) multi-stage architecture \cite{chen2019hybrid} mainly follows the idea that a single detector is unlikely to be able to train uniformly on all quality levels of IoU.
The cascade architecture tries to solve the EVPS problem by training multiple regressors connected sequentially, each of which is specialized in a predetermined and growing IoU minimum threshold.
Each regressor performs a conversion of its localization results into a new list of proposals for the following regressor.
Although this type of architecture clearly improves the overall performance, it also introduces a considerable number of new parameters into the network.
In fact, with respect to its predecessor Mask R-CNN, the number of detection and segmentation modules triples.
To reduce the complexity of cascade networks and to address the EVPS problem, we design a lighter architecture with single detection and mask heads uniformly trained on all IoU levels.
Authors in \cite{Cai_2018_CVPR} underlined the cost-sensitive learning problem \cite{elkan2001foundations, masnadi2010cost}, where the optimization of different IoU thresholds requires various loss functions.
Inspired by this study, we address the problem using multiple selective training, which focuses on a specific IoU quality in each step and recursively feeds them into the detector.
The intuition is that the detector training and its ability to return an adequate number of proposals of a certain quality level will happen at the same time.

In Fig. \ref{fig:recref-arch} (b), the new $R^3$-CNN architecture along with our training paradigm are shown.
In this loop (recursive) architecture, the detector and the RoI pooling modules are connected in a cycle.
As in HTC, the first set of RoI proposals is provided by the RPN.
After that, the RoI pooler crops and converts them to fixed-size feature maps, which are used to train the B1 block.
Then, with an appropriate IoU threshold, the ground truth re-sampling takes place by the B1 block to generate a new proposal set.
The result is then used both in the segmentation module M1 and as the new input for the pooler which closes the loop.
By the IoU threshold manipulation, the network can force the detection to extract those RoIs with IoU quality levels which are typically missed.
The cycle continues three times (3x loop) to guarantee the rebalancing of RoI levels.

\begin{figure}[t]
	\begin{center}
		\includegraphics[trim=0 0 0 0, clip, width=1\linewidth]{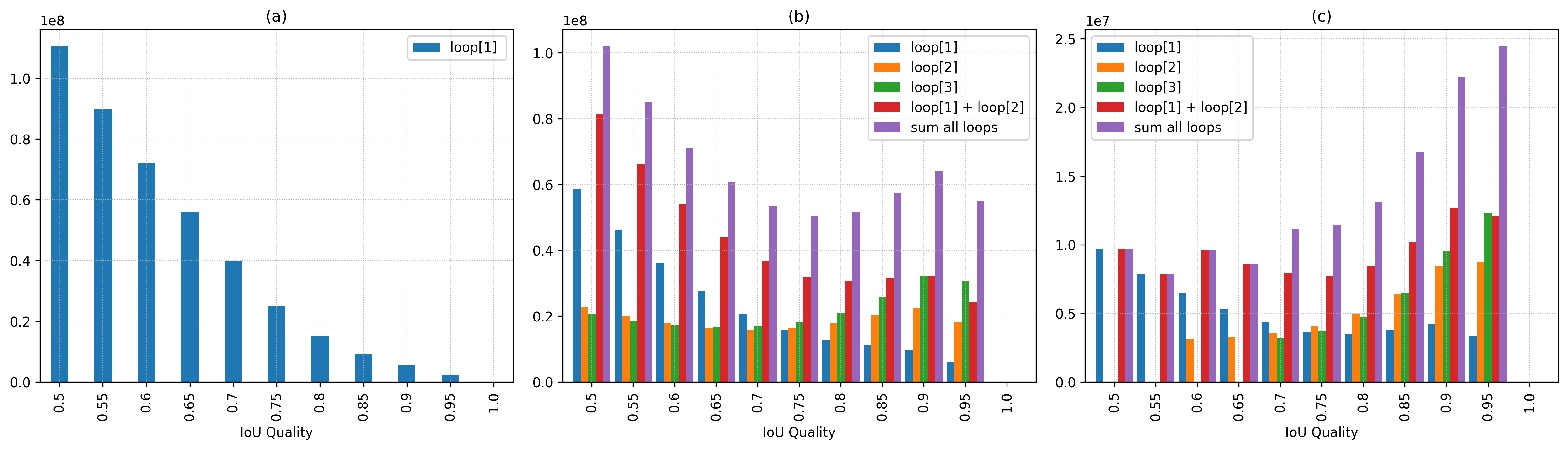}
	\end{center}
	\caption{The IoU histogram of training samples for Mask R-CNN with a 3x schedule (36 epochs) (a), and $R^3$-CNN where each loop uses different IoU thresholds [0.5, 0.6, 0.7], decreasingly (b) and increasingly (c). Better seen in color.}
	\label{fig:iou-histogram}
\end{figure}

Fig. \ref{fig:iou-histogram} (a) shows the generated RoI distribution for each IoU level in Mask R-CNN as well as the EVPS problem.
The distribution of the rebalanced samples by our model, on the other hand, can be seen in Fig. \ref{fig:iou-histogram} (b) and (c).
For the latter, it is worth emphasizing some important details emerging from these graphs: (i) Considering only the first loop trend, $R^3$-CNN looks quite similar to Mask R-CNN; (ii) Conversely, considering the sum of the first two loops, our distribution looks much more balanced; (iii) The third loop significantly increases the number of high-quality RoIs.
Despite the fact that our architecture contains a single detector, its behavior shows a unique and well-defined trend in terms of RoI distribution within different loops.
We believe this is the reason why $R^3$-CNN outperforms Mask R-CNN.
It is able to mimic the Cascade R-CNN behavior in the RoI distribution (as also shown in \cite{Cai_2018_CVPR}), achieved by HTC, but using only a single detector and significantly fewer parameters.

For a given loop $t$, let us define $h$ as the sole classifier and $f$ as the sole regressor which is trained for a selected IoU threshold $u^t$, with $u^t > u^{t-1}$, by minimizing the loss function of Cascade R-CNN \cite{Cai_2018_CVPR}:
\begin{equation}
L(x^t, g) = L_{cls}\left(h\left(x^t\right), y^t\right) + \lambda\left[y^t \geqslant 1\right]L_{loc}\left(f\left(x^t, b^t\right), g\right)
\end{equation}

\noindent where $x^t$ represents the input features of the $t$-th loop, $b^t = f\left(x^{t-1}, b^{t-1}\right)$ is the new sampled set of proposals coming from the previous loop (with $b^0$ coming from the RPN), $g$ is the ground truth, and $\lambda$ is a positive coefficient.
$y^t$ represents the label of $x^t$ given the IoU threshold $u^t$, the proposals $b^t$ and the ground truth label $g_y$ with the following equation:
\begin{equation}
y^t = \begin{cases}
  g_y   & \text{if  } IoU\left(b^t, g\right) \geqslant u^t  \\
  0     & \text{otherwise} \\
\end{cases}
\end{equation}

At inference time, the same loop procedure is applied and all the predictions are merged together by computing the mean of the classification values.
As it will be shown in the experiments, using loops also at inference (or evaluation) time is not optional, meaning that the loop mechanism is intrinsic to the weights of the network.

\section{Experiments}
\label{sec:experiments}
\subsection{Dataset and Evaluation Metrics}
\label{subsec:dataset}
\noindent\textbf{Dataset}.
As the majority of recent literature on instance segmentation, we perform our tests on the MS COCO 2017 dataset \cite{lin2014microsoft}.
The training dataset consists of more than 117,000 images and 80 different classes of objects.

\noindent\textbf{Evaluation Metrics}.
We used the same evaluation functions offered by the python \textit{pycocotools} software package.
All the evaluation phases have been performed on the COCO minival 2017 validation dataset, which contains 5000 images. We report the Average Precision (AP) with different IoU thresholds for both bounding box and segmentation tasks.
The main metric ($AP$) is computed with IoUs from 0.5 to 0.95.
Others include $AP_{50}$ and $AP_{75}$ with 0.5 and 0.75 minimum IoU thresholds, and $AP_s$, $AP_m$ and $AP_l$ for small, medium and large objects, respectively.

\subsection{Implementation details}
To perform a fair comparison, we obtain all the reported results by training the networks with the same hardware and, when possible, the same software configuration.
When available, the original code released by the authors or the corresponding implementation in MMDetection \cite{mmdetection} framework were used.
Our code is also developed within this framework.
In the case of HTC, we did not consider the semantic segmentation branch.

We performed a distributed training on 2 servers, each equipped with 2x16 IBM POWER9 cores, 256 GB of memory and 4 x NVIDIA Volta V100 GPUs with Nvlink 2.0 and 16GB of memory.
Each training consists of 12 epochs with Stochastic Gradient Descent (SGD) optimization algorithm, an initial learning rate of 0.02, a weight decay of 0.0001, and a momentum of 0.9. 
The learning rate decays at epochs 8 and 11. 
We used batch size of 2 for each GPU.
We fixed the long edge and short edge of the images to 1333 and 800, maintaining the aspect ratio.
ResNet 50 \cite{he2016deep} was used as the backbone.
If not specified differently, the number of loops in training and evaluation are the same.
\subsection{Analysis of $R^3$-CNN}
\noindent\textbf{Description}.
In this part, we demonstrate the potentiality offered by a naive three-stage loop compared to Mask R-CNN and the original three-stage cascade HTC.
To have a fair comparison, we select the optimal configuration for the HTC network as baseline and also apply it to training our $R^3$-CNN. 
In the \textit{advanced} version, we replace fully-connected layers from detection head with lightweight convolutions with kernel $7\times7$ and a Non-Local block \cite{wang2018non} with incremented kernel size of $7 \times 7$ to better exploit information.
We also build a brand new branch using only convolutions and Non-Local blocks to include a new learning task to improve segmentation as described in \cite{huang2019mask}.
Since our naive version has slightly fewer parameters than Mask R-CNN, it is also insightful to compare it with our model.
Finally, we also want to demonstrate the following important claim: it does not matter the way or order with which the IoU thresholds are changed (either incrementally or decrementally), since in both cases a more balanced IoU distribution is achieved (see Figure \ref{fig:iou-histogram}).

\noindent\textbf{Results}.
In Table \ref{tab:count_parameters}, we report speed in evaluation and memory usage in training, distinguishing between \textit{memory usage} of the entire training process and \textit{model size} (proportional to the number of parameters). 
Comparing the naive version (row \#4) with HTC (row \#3), it can be seen that our model has significantly fewer parameters and is more memory efficient.
While the segmentation precision ($S_{AP}$) is practically the same, there is a slight loss in $B_{AP}$.
Also, the speed of naive is slightly better than HTC.
Regarding the advanced version (row \#6), it surpasses the HTC accuracy in both tasks, while saving a significant number of parameters and using the same amount of memory in training.
The only disadvantage is the reduced speed due to Non-Local blocks.

Compared to Mask R-CNN (row \#1), the naive $R^3$-CNN has the same complexity, but achieves a much higher precision in both tasks.
To further investigate how well our recursive mechanism works, we also compare it with Mask R-CNN trained with triple number of epochs (row \#2).
While more training helps Mask R-CNN produce a higher precision, it is still outperformed by naive $R^3$-CNN. This demonstrates that our loop mechanism is not simply another way of training the network for more epochs, but that it represents a different and more effective training strategy. This can be explained by the fact that while in Mask R-CNN the RoI proposals are always provided by the RPN, in our case they are provided by the detection head which generates higher quality and more balanced RoIs (see Figure \ref{fig:iou-histogram}).

Finally, to show that the order of changes in IoU threshold is not crucial to performance, in rows \#4 and \#5, we report a comparison between increasing and decreasing IoU thresholds through loops.
Although there is a slight degradation of precision using the decreasing training strategy, it is almost negligible due to a more balanced IoU distribution achieved in both cases, but skewed to high-quality RoIs in the first case and low-quality in the latter.
\begin{table}[t]
	\small
	\setlength{\tabcolsep}{2.5pt}
	\begin{center}
		\begin{tabular}{c|l|c|c|c|c|c|c|c|c|c}
			\# & Model & \# Params & TS & $L_t$ & $H$ & $B_{AP}$ & $S_{AP}$ & Speed & Mem. usage & Model size \\
			\hline\hline
			1 & Mask (1x)           & 44,170 K & -   & 1 & 1 & 38.2 & 34.7 & 11.5 & 4.4 GB  & 339 MB \\
			2 & Mask (3x)           & 44,170 K & -   & 1 & 1 & 39.2 & 35.5 &  5.4 & 4.4 GB  & 339 MB \\
			3 & HTC                 & 77,230 K & Inc & 3 & 3 & 41.7 & 36.9 &  5.4 & 6.8 GB  & 591 MB \\
			\hline
			4 & $R^3$-CNN (naive)   & 43,912 K & Inc & 3 & 1 & 40.9 & 36.8 &  5.5 & 5.9 GB  & 337 MB \\
			5 & $R^3$-CNN (naive)   & 43,912 K & Dec & 3 & 1 & 40.4 & 36.7 &  5.5 & 5.9 GB  & 337 MB \\
			6 & $R^3$-CNN (advanced)& 50,072 K & Inc & 3 & 1 & 42.0 & 38.2 &  1.0 & 6.8 GB  & 384 MB
		\end{tabular}
	\end{center}
	\caption{Comparing trainable parameters with the bounding box and segmentation average precision. K: thousand. Column $L_t$: number of stages. Speed is image per second.
		TS: Training strategies. Inc: progressively increasing and Dec: decreasing IoU quality through loops.}
	\label{tab:count_parameters}
\end{table}

\subsection{Ablation study on the evaluation phase}
\noindent\textbf{Description}.
In this subsection we focus on how the results are affected by the number of cycles in the evaluation phase.
We consider the naive version mentioned above as pre-trained model, which consists of three loops in the training and evaluation phases.

\noindent\textbf{Results}.
From Table \ref{tab:compare-num-eval-stages}, it is evident that the loop mechanism is of paramount importance for the evaluation phase too.
In fact, when we train the network with the 3x loops and then evaluate it with one loop, it performs even worse than Mask R-CNN (row \#2).
On the contrary, with two loops, the result is significantly better (row \#3).
It also underperforms the three-loop evaluation only slightly (row \#4). 
Therefore, this version could be considered a good compromise between execution time and detection quality.
From four loops onward, the performance tends to remain almost stable.
This is consistent with our initial hypothesis of a link between evaluation and the loop mechanism, and confirms that in order to have higher performance with more than three loops in the evaluation, we also need to increase the number of loops in the training phase.
\subsection{Ablation study on the training phase}

\noindent\textbf{Description}.
In this experiment, the network is trained with a number of loops varying from 1 to 4.
The number of loops for the evaluation changes accordingly.

\noindent\textbf{Results}.
The comparative results are reported in Table \ref{tab:compare-num-training-stages}.
The precision of the single loop (row \#2) is comparable to Mask R-CNN, and not much different from the above-mentioned model with one-loop evaluation (row \#2 of Table \ref{tab:compare-num-eval-stages}).
This connection with the previous experiment suggests that the detector is strictly optimized on the corresponding sample distribution.
The performance is improved by the training strategies with two and three loops, though significantly by the former and only slightly over that by the latter. 
Regarding more than 3 loops (row \#5), the improvement is negligible.
\begin{table}[t]
	\centering
	\makebox[0pt][c]{\parbox{1\textwidth}{
			\begin{minipage}[c]{0.48\textwidth}\centering
				\begin{center}
					\begin{tabular}{c|l|c|c|c|c|c}
						\# & Model & $L_t$ & $H$ & $L_e$ & $B_{AP}$ & $S_{AP}$ \\
						\hline\hline
						1 & Mask       & 1 & 1 & 1 & 38.2 & 34.7  \\
						\hline
						2 & \multirow{5}{*}{$R^3$-CNN}     & 3 & 1 & 1 & 37.9 & 35.4 \\
						3 &          & 3 & 1 & 2 & 40.5 & 36.6 \\
						4 &          & 3 & 1 & 3 & 40.9 & 36.8 \\
						5 &          & 3 & 1 & 4 & 40.9 & 36.7 \\
						6 &          & 3 & 1 & 5 & 40.9 & 36.7
					\end{tabular}
				\end{center}
				\caption{Impact of evaluation loops $L_e$ in a 3-loop and one-head-per-type $R^3$-CNN model. Row \#4 is the naive $R^3$-CNN in Table \ref{tab:count_parameters}.}
				\label{tab:compare-num-eval-stages}
			\end{minipage}
			\hfill
			\begin{minipage}[c]{0.48\textwidth}\centering
				\begin{center}
					\begin{tabular}{c|l|c|c|c|c}
						\# & Model & $L_t$ & $H$ & $B_{AP}$ & $S_{AP}$ \\
						\hline\hline
						1 & Mask & 1 & 1 & 38.2 & 34.7 \\
						\hline
						2 & \multirow{4}{*}{$R^3$-CNN}     & 1 & 1 & 37.7 & 34.7 \\
						3 &           & 2 & 1 & 40.4 & 36.4 \\
						4 &           & 3 & 1 & 40.9 & 36.8 \\
						5 &           & 4 & 1 & 40.9 & 36.8
					\end{tabular}
				\end{center}
				\caption{Impact of the number of training loops in a one-head-per-type $R^3$-CNN model. Row \#4 is the naive $R^3$-CNN in Table \ref{tab:count_parameters}.}
				\label{tab:compare-num-training-stages}
			\end{minipage}
	}}
\end{table}
\subsection{Extensions on $R^3$-CNN}


\noindent\textbf{Description}.
Our final experiments show that $R^3$-CNN model can be plugged in seamlessly to several state-of-the-art architectures for instance segmentation, consistently improving their performance, which demonstrates its generalizability.
In this experiment, we select our best-performing version previously called \textit{advanced} (see Table \ref{tab:count_parameters} row \#6) and renamed $R^3$-CNN-L model.
The experiment tested different state-of-the-art models, namely GRoIE \cite{rossi2020novel}, GC-net \cite{cao2019gcnet}, DCN \cite{zhu2019deformable}, with and without the $R^3$-CNN-L version.
When compatible, we also merged the original model with HTC as baseline.
For example, \emph{HTC+GC-Net} is composed of both HTC and GC-Net merged together.

\noindent\textbf{Results}.
The results are summarized in Table \ref{sota-coco-bbox-mask-results}, where best results for each comparison are reported in bold, while the second best is in red.
From the table we can see that almost all the best and second-best scores belong to $R^3$-CNN architectures, both in object detection and instance segmentation.
However, there are two cases in which this does not happen. 
The first one is the combination of GC-Net and HTC which outperforms $R^3$-CNN-L in $AP_m$ by $0.6\%$ (row \#7), and the other one is the combination of DCN and HTC which outperforms $R^3$-CNN-L in $AP_l$ by the same amount (row \#10). 
Apart from these rare cases, these experiments confirm that the proposed $R^3$-CNN consistently brings benefits to existing object detection and instance segmentation models, in terms of both precision and reduced number of parameters.

\begin{table*}[t]
	\footnotesize
	\setlength{\tabcolsep}{1.2pt}
	\begin{center}
		\begin{tabular}{c|lH||c|c|c|c|c|c||c|c|c|c|c|c}
			& & & \multicolumn{6}{c||}{Bounding Box (object detection)} & \multicolumn{6}{c}{Mask (instance segmentation)} \\
			\# & Method & Backbone & $AP$ & $AP_{50}$ & $AP_{75}$  & $AP_{s}$ & $AP_{m}$ & $AP_{l}$  & AP & $AP_{50}$ & $AP_{75}$  & $AP_{s}$ & $AP_{m}$ & $AP_{l}$\\
			\hline\hline
			1 & Mask                 & r50-FPN  & 37.3 & 58.9 & 40.4 & 21.7 & 41.1 & 48.2 & 34.1 & 55.5 & 36.1 & 18.0 & 37.6 & 46.7 \\
			2 & HTC                  & r50-FPN  & \red{41.7} & \red{60.4} & \red{45.2} & \red{24.0} & \red{44.8} & \red{54.7} & \red{36.9} & \red{57.6} & \red{39.9} & \red{19.8} & \red{39.8} & \red{50.1} \\
			3 & $R^3$-CNN-L             & r50-FPN  & \one{42.0} & \one{61.0} & \one{46.3} & \one{24.5} & \one{45.2} & \one{55.7} & \one{38.2} & \one{58.0} & \one{41.4} & \one{20.4} & \one{41.0} & \one{52.8} \\
			\hline\hline
			4 & GRoIE                & r50-FPN  & \red{38.6} & \red{59.4} & \red{42.1} & \red{22.5} & \red{42.0} & \red{50.5} & \red{35.8} & \red{56.5} & \red{38.4} & \red{19.2} & \red{39.0} & \red{48.7} \\
			5 & $R^3$-CNN-L+GRoIE       & r50-FPN  & \one{42.0} & \one{61.2} & \one{45.6} & \one{24.4} & \one{45.2} & \one{55.7} & \one{39.1} & \one{58.8} & \one{42.3} & \one{20.7} & \one{42.1} & \one{54.3} \\
			\hline\hline
			6 & GC-Net               & r50-FPN  & 40.5 & 62.0 & 44.0 & 23.8 & 44.4 & 52.7 & 36.4 & 58.7 & 38.5 & 19.7 & 40.2 & 49.1 \\
			7 & HTC+GC-Net           & r50-FPN  & \red{43.9} & \red{63.1} & \red{47.7} & \red{26.2} & \one{47.7} & \red{57.6} & \red{38.7} & \red{60.4} & \red{41.7} & \red{21.6} & \red{42.2} & \red{52.5} \\
			8 & $R^3$-CNN-L+GC-Net      & r50-FPN  & \one{44.3} & \one{64.1} & \one{48.4} & \one{27.0} & \red{47.1} & \one{58.9} & \one{40.2} & \one{61.1} & \one{43.5} & \one{22.6} & \one{42.8} & \one{56.0} \\
			\hline\hline
			9 & DCN                 & r50-FPN  & 41.9 & 62.9 & 45.9 & 24.2 & 45.5 & 55.5 & 37.6 & 60.0 & 40.0 & 20.2 & 40.8 & 51.6 \\
			10 & HTC+DCN             & r50-FPN  & \red{44.7} & \red{63.8} & \red{48.6} & \red{26.5} & \red{48.2} & \one{60.2} & \red{39.4} & \red{61.2} & \red{42.3} & \red{21.9} & \red{42.7} & \red{54.9} \\
			11 & $R^3$-CNN-L+DCN        & r50-FPN  & \one{44.8} & \one{64.3} & \one{48.9} & \one{26.6} & \one{48.3} & \red{59.6} & \one{40.4} & \one{61.3} & \one{44.0} & \one{22.3} & \one{43.6} & \one{56.1} \\
		\end{tabular}
	\end{center}
	\caption{Performance of the state-of-the-art models with and without $R^3$-CNN model. Bold values are best results, red ones are second-best values.}
	\label{sota-coco-bbox-mask-results}
\end{table*}

\section{Conclusions}
\label{sec:conclusion}

In this paper, we introduced the $R^3$-CNN architecture to address the issue of exponentially vanishing positive samples in training by rebalancing the training proposals with respect to the IoU thresholds, through a recursive re-sampling mechanism in a single detector architecture.
We demonstrated that a good training needs to take into account the diversity of IoU quality of the RoIs used to learn, more than aiming to have only high quality RoIs.
Our extensive set of experiments and ablation studies provide a comprehensive understanding of the benefits and limitations of the proposed models.
$R^3$-CNN offers a good flexibility to use intermediate versions between the naive version and HTC, permitting to play with the number of loops, depending if we privilege precision, number of parameters or speed.
Overall, the proposed $R^3$-CNN architecture demonstrates its usefulness when used in conjunction with several state-of-the-art models, achieving considerable improvements over the existing models.

%
%
%
 \bibliographystyle{splncs04}
 \bibliography{paper}

\end{document}